\documentclass[10pt,twocolumn,letterpaper]{article}

\usepackage{iccv}
\usepackage{times}
\usepackage{epsfig}
\usepackage{graphicx}
\usepackage{amsmath}
\usepackage{amssymb}
\usepackage{booktabs}
\usepackage{setspace}
\usepackage{multirow}
\usepackage{makecell}
\usepackage {mathtools}
\usepackage{amsmath}
\usepackage{amssymb}
\usepackage{bbm}
\usepackage{dsfont}
\usepackage{subfigure}
\usepackage[accsupp]{axessibility}
\input{def.set}
\usepackage[breaklinks=true,bookmarks=false]{hyperref}

\iccvfinalcopy 

\def\iccvPaperID{****} 

\ificcvfinal\fi

\begin{document}

\title{A Style and Semantic Memory Mechanism for Domain Generalization\thanks{{\small This work was performed at JD AI Research.}}}
\author{Yang Chen$^\dagger$, Yu Wang$^\ddagger$, Yingwei Pan$^\ddagger$, Ting Yao$^\ddagger$, Xinmei Tian$^\dagger$, and Tao Mei$^\ddagger$\\
\small $^\dagger$ University of Science and Technology of China, Hefei, China\\
\small $^\ddagger$JD AI Research, Beijing, China
\\
{\tt\scriptsize cheny01@mail.ustc.edu.cn, \{feather1014, panyw.ustc, tingyao.ustc\}@gmail.com, xinmei@ustc.edu.cn, tmei@jd.com}
}

\maketitle
\ificcvfinal\thispagestyle{empty}\fi

\begin{abstract}
Mainstream state-of-the-art domain generalization algorithms tend to prioritize the assumption on semantic invariance across domains. Meanwhile, the inherent intra-domain style invariance is usually underappreciated and put on the shelf. In this paper, we reveal that leveraging intra-domain style invariance is also of pivotal importance in improving the efficiency of domain generalization. We verify that it is critical for the network to be informative on what domain features are invariant and shared among instances, so that the network sharpens its understanding and improves its semantic discriminative ability. Correspondingly, we also propose a novel ``jury'' mechanism, which is particularly effective in learning useful semantic feature commonalities among domains. Our complete model called STEAM can be interpreted as a novel probabilistic graphical model, for which the implementation requires convenient constructions of two kinds of memory banks: semantic feature bank and style feature bank. Empirical results show that our proposed framework surpasses the state-of-the-art methods by clear margins.
\end{abstract}

\section{Introduction}
Machine learning models are usually deployed in scenarios where the test data are unknown beforehand. This phenomenon can lead to dangerous consequences especially when the predictions are used for life-threatening occasions such as medical diagnosis. The prediction might be seriously erroneous due to the distributional gap between training data and test data. It is therefore critical for machine learning algorithms to maintain safe and reliable predictions that generalize well across domains. The goal of domain generalization (DG) approaches \cite{li2018domain, li2018deep, shankar2018generalizing, zhou2021domain} is to solve this issue by leveraging labeled data from multiple training domains. However, we observed that mainstream state-of-the-art DG algorithms tend to only prioritize the semantic invariance assumption across domains, while the style invariance within each domain is usually ignored. In this paper, we reveal that intra-domain style invariance is also of pivotal importance to improve DG approaches. Particularly, we propose a novel model to incorporate both intra-domain style invariance and inter-domain semantic invariance for DG tasks. The proposed framework is called \textbf{STEAM}, which relies on a \textbf{ST}yle and s\textbf{E}m\textbf{A}ntic \textbf{M}emory mechanism to practically implement our proposed assumptions.

In contrast to existing DG works \cite{carlucci2019domain, zhou2020learning, zhou2021domain}, STEAM further benefits from the hypothesis that instances from the same domain should share style information. The motivation is that the simple constraint helps efficiently disentangle the style feature, and therefore eases the search for true semantic feature with a reduced degree of freedom. To reach this goal, we resort to the recently prevailing self-supervised learning paradigm that makes our assumptions practically accessible. Our first objective is to achieve intra-domain style invariance by conveniently resorting to contrastive loss. Given the invariance assumption, style features corresponding to each domain are discovered, and the network can further learn semantic features along the directions mostly orthogonal to the domain styles. Since the semantic feature is considered as the true causal factor affecting the instance category, STEAM effectively helps reduce overfitting to the domain styles through the above mechanisms. Most importantly, we also force the network to respect the conventional semantic invariance among domains. Specifically, we require that each pair of samples from the same class to compute a similarity score with all the semantic features in ``memory''. These two samples need to reach a consensus on every such similarity score when sweeping through all the stored semantic features. We name this procedure ``jury'' mechanism, and we elaborate the mathematical net effect of such mechanism in the paper.

To summarize, our contributions in this paper include: 1. We explore the assumption of instance level intra-domain style invariance and justify the empirical advantage by incorporating such assumptions into the DG framework. 2. We propose a ``jury'' mechanism, which efficiently learns domain-agnostic semantic features beneficial for classification tasks. Such mechanism is capable of efficiently preventing semantic features from overfitting to domain styles. 3. We observe that the proposed STEAM not only generalizes well on DG benchmarks, but also can be conveniently modified for domain adaptation (DA) problems.

\section{Related Work}
\textbf{Domain Adaptation} (DA) algorithms aim to exploit both annotated training data in the source domain and unlabeled samples in the target domain. Mainstream DA approaches  \cite{cai2019exploring,chen2019mocycle,pan2019transferrable,pan2020exploring,yao2015semi} usually penalize distributional misalignment between source and target data via, e.g., maximum mean discrepancy loss \cite{long2015learning,long2017deep} or adversarial loss \cite{bousmalis2017unsupervised, long2018conditional, sankaranarayanan2018generate}. Given the success of DA algorithms, multi-source domain adaptation (MSDA) methods \cite{peng2019moment, xu2018deep, zhao2018adversarial} consider scenarios where multiple sources are available for training to better improve generalization.

\textbf{Domain Generalization} (DG) algorithms also assume access to multiple labeled training source domains. However, target data is unavailable during training for DG approaches, leading to a more challenging yet a more practical setting than DA problems. Many early DG approaches \cite{ li2018domain, li2018deep, motiian2017unified} borrowed the idea of distribution alignment from DA to reduce the distributional gap between multiple training sources. Some recent DG methods consider generating extra synthetic images given the multiple source domains, so that test data distributed closely to the training data are ``in-distribution'' with the training data \cite{bai2020decaug, shankar2018generalizing, somavarapu2020frustratingly, zhou2020learning}. Some methods decompose the network parameters into domain-specific and domain-invariant parts during training, while only the domain-invariant parameters are used for predictions at test time. For instance, \cite{li2017deeper} develops a low-rank parameterized CNN model where each layer of the network is decomposed into ``common'' and ``specific'' components. In \cite{piratla2020efficient}, only the last layer of the network is decomposed to serve the goal of DG. Several normalization and meta-learning strategies are also considered for domain generalization such as \cite{li2019episodic, seo2019learning, zhou2021domain}.

\textbf{Contrastive Learning} has shown impressive performance in the context of self-supervised learning \cite{cai2020joint, chen2020simple, he2020momentum,mitrovic2020representation,wang2021low,wu2018unsupervised,yao2021seco}. Representative contrastive learning \cite{he2020momentum} proposed efficient memory bank constructions so that the historical features are conveniently stored and reused even if batchsize is small. We feel inspired from these approaches \cite{he2020momentum} where the memory bank help retain temporal consistency between features, and we introduce such a memory mechanism to best facilitate our motivations. However, our approach is neither targeting a pre-train task nor an unsupervised learning problem, in contrast to \cite{chen2020simple, he2020momentum, wu2018unsupervised}. Rather, we look into DG problems and we aim to learn the desired feature invariances respecting our unique hypothesis. We notice a related contrastive learning method in \cite{mitrovic2020representation}, that enforces a particular prediction regularizer across augmentations to improve in-class consistency. We argue that the work in \cite{mitrovic2020representation} is an unsupervised algorithm that completely distinguishes its nature from our DG task, while the loss proposed here also leads to an entirely different interpretation and application.

We observe that existing DG methods along the decomposition path either: hinge on inter-domain semantic invariance features assumption, so that the network becomes agnostic to styles; or they rely on style decomposition approaches to synthesize more data. In contrast, our proposed method enjoys two exclusive novel assumptions: 1. We are the first to impose instance level intra-domain style invariance during the training. Having these domain-specific style features at hand, we effectively reduce the degree of freedom of the problem in further learning the useful semantic features.  2. We design a novel ``jury'' mechanism that is different from any existing DG method, which achieves significant improvement in domain generalization.

\section{Method}
In the regime of out of distribution (OOD) detection, it has been shown that data distribution is heavily affected by population-level background statistics \cite{choi2018generative,nalisnick2018deep,ren2019likelihood,serra2019input}. Owing to this issue, OOD inputs can rather be classified into in-distribution classes with high confidence, given the presence of dominant background noise. Recent observations \cite{choi2018generative,nalisnick2018deep} have shown that deep generative models can even assign a higher likelihood to OOD inputs. One reason is that simple parameterization on marginal input distribution can be significantly confounded and dominated by background statistics, and does not learn much useful parameterization on the variation of {\emph{semantic features}}. Accordingly, work such as \cite{ren2019likelihood} aims to mute the effect of these background statistics via the likelihood ratio method, to achieve better OOD detection based on cleaner semantic features. Although the literature on OOD detection investigates completely orthogonal topics and directions against the DG community, we feel intrigued and motivated to reformulate domain generalization problems with novel background noise priors. The ultimate goal is to relieve the network optimization on the background statistics, so that the network steers away from its overfitting to these domain specific background noise, while focusing on learning true semantic distributional commonalities among domains.

We therefore do not seek to directly close the distribution gap between data distributions across domains (that GAN and MMD methods normally do). We impose prior knowledge that {\emph{ instances from the same domain share hidden invariant background statistics}}. Under this constraint, the network reduces uncertainty and redundancy when searching for semantic features that might be glimmering in comparison to the dominant background style statistics.

\begin{figure}[t]
\center
\includegraphics[width=0.8\linewidth]{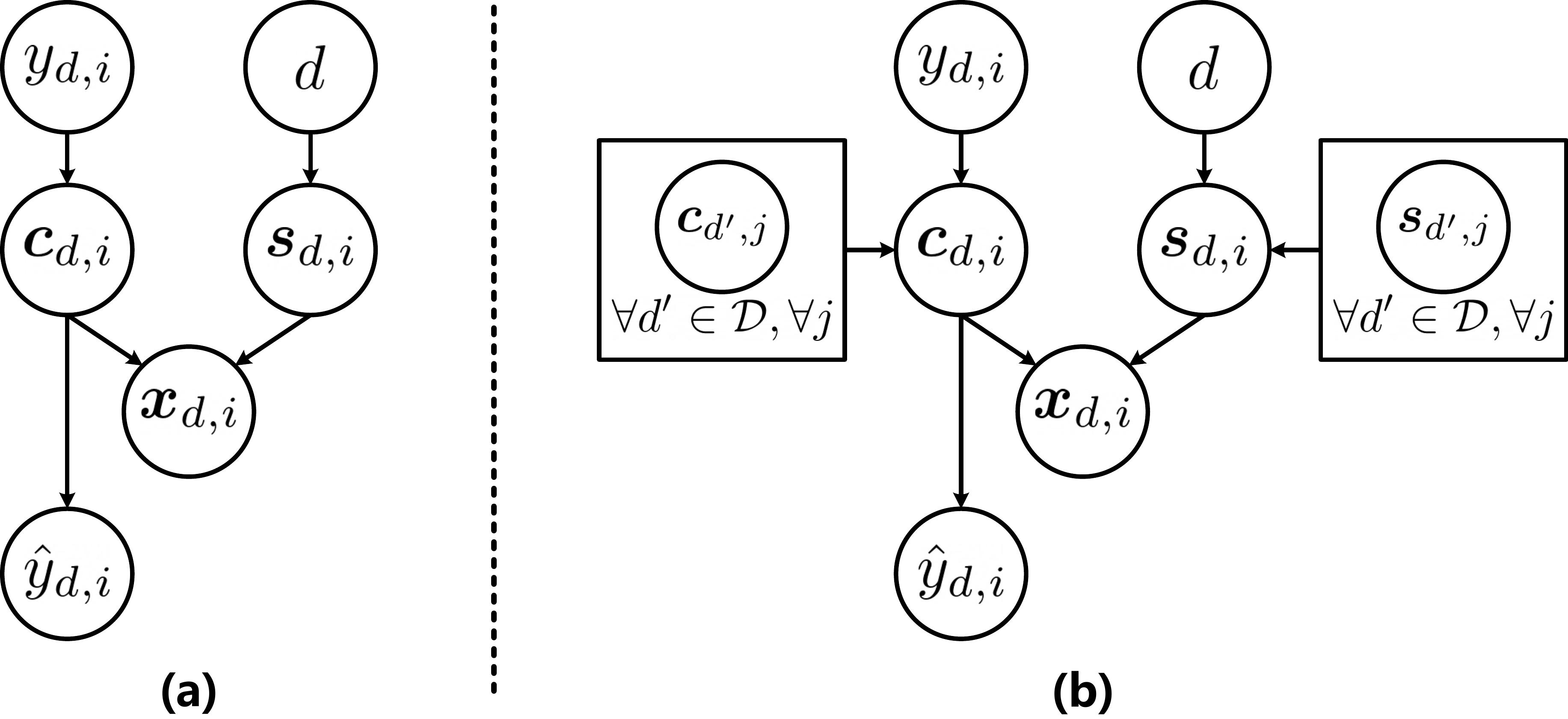}
\caption{Probabilistic graphical model comparisons for (a) conventional causal model for DG (b) causal model for STEAM, based on memory bank constructions. For STEAM, all the training samples have an impact on deciding the semantic and style feature for each instance. $\hat{y}_{d,i}$ is the class prediction given feature $\bc_{d,i}$.}
\label{fig:causal_graph}
\end{figure}

\subsection{Problem Formulation} \label{sec:probform}
For DG problems, we consider $D$ source domains $\mathcal{D}=\{\mathcal{D}_{d}\}^{D}_{d=1}$, with each $d$-th domain $\mathcal{D}_{d}$ includes $N_{d}$ training pairs $\{(\bx_{d,i},y_{d,i})\}^{N_{d}}_{i=1}$, where $\bx_{d,i}$ is the $i$-th sample in $\mathcal{D}_{d}$, and $y_{d,i} \in \{1,2,...,n_c\}$ is the label of $\bx_{d,i}$. $n_c$ is the number of classes shared across domains. The goal of DG approach is to learn a model from multiple labeled source domains that generalizes well to an unseen target domain $D_T$.

We define notations frequently used throughout the paper. We assume each training sample is projected to feature embedding via CNN encoders \cite{he2016deep, li2021contextual}. Specifically, the image $\bx_{d,i}$ firstly is input into feature extractor: $\bz_{d,i}=E_{f}(\bx_{d,i}, \btheta_{e, f})$, where function $E_{f}$ extracts image feature $\bz_{d,i}$ out of $\bx_{d,i}$ by a CNN parameterization $\btheta_{e, f}$. A semantic encoder $E_{c}$ then reads in $\bz_{d,i}$ and produces semantic feature $\bc_{d,i}= E_{c}(\bz_{d,i}, \btheta_{e, c})$. In parallel, a style encoder simultaneously inputs the $\bz_{d,i}$ feature and maps it into style representation via $\bs_{d,i}=E_{s}(\bz_{d,i}, \btheta_{e, s})$. We define a classifier $C(\bc_{d,i}, \bphi)$, where function $C$ is parameterized by $\bphi$ and used to classify semantic features among $n_c$ possible classes. Here, the above encoder parameters $\bTheta_{e}=\{\btheta_{e, f},\btheta_{e, c},\btheta_{e, s}\}$ and classifier parameters $\bphi$ are CNN parameters learned during training. Subscript $e$ associated with each notation, e.g.,  $\btheta_{e, c}$, is intended to be reminiscent of ``encoder''.

We name the proposed method ``Style and Semantic Memory Mechanism (STEAM)'' for Domain Generalization. The overall STEAM model can be interpreted as a probabilistic graphical model in Fig. \ref{fig:causal_graph}(b). Each semantic feature $\bc_{d,i}$ and style feature $\bs_{d,i}$ both depend on statistics (rectangles) from every other instances available in memory (given unlimited memory, the whole dataset then). Our motivation is that human intrinsically define classes by contrasting. Take for instance, Labrador and Husky are both defined as dogs, apparently because they share higher similarity than with any other species. In comparison, conventional DG methods as shown as in Fig. \ref{fig:causal_graph}(a) only models each instance's semantic feature $\bc_{d,i}$ independently from other $\bc_{d',j}$. Note STEAM hinges on enormous historical training data. It is therefore critical that our loss function breaks free from the limitation of batchsize. A simple solution would be to directly store the learned features out of the encoder into a static memory bank for later usage \cite{wu2018unsupervised}. Unfortunately, similar to the observation in \cite{he2020momentum}, we find features stored in this way cannot retain any temporal representation consistency due to the rapidly update of encoder via backpropagation. We thus simultaneously maintain an alternative memory encoder: $E_m=\{E_{m,f}, E_{m,c}, E_{m,s}\}$ with parameters $\bTheta_{m}=\{\btheta_{m, f},\btheta_{m, c},\btheta_{m, s}\}$ that is able to slowly release the historical feature representations into the memory bank in an momentum updated way.

Specifically, we make sure the constructed memory encoder shall involve identical architecture and functioning components mirroring everything in $E_{c}, E_{f}, E_{s}$ above: We have memory feature encoder: $E_{m,f}(\cdot, \btheta_{m, f})$, memory style feature encoder $E_{m, s}(\cdot, \btheta_{m, s})$, and memory semantic feature encoder $E_{m,c}(\cdot, \btheta_{m, c})$. The only difference is that the parameters $\bTheta_{e}=\{\btheta_{e, f},\btheta_{e, c},\btheta_{e, s}\}$  are updated via backpropagation, whereas the parameters of memory encoder $\bTheta_{m}=\{\btheta_{m, f},\btheta_{m, c},\btheta_{m, s}\}$ are only momentum updated according to the changes of $\bTheta_e$. Subscript $m$ is intended to be reminiscent of ``memory''. We elaborate the usage of memory encoders in Section \ref{sec:intrastyle}, and Section \ref{sec:intersemantic}.

\begin{figure*}[th!]
\begin{center}
\vspace{-0.2in}
\includegraphics[width=0.83\linewidth]{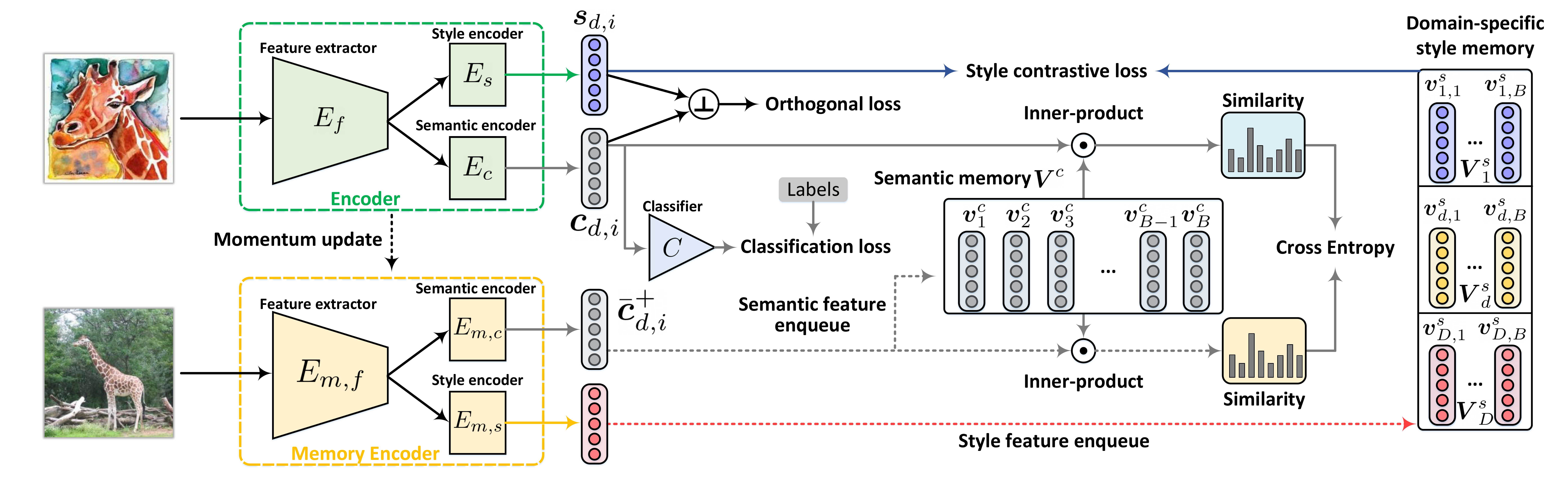}
\end{center}
\vspace{-0.2in}
\caption{The framework of STEAM. We train an encoder for style and semantic feature extraction. We maintain a memory encoder to obtain $D$ number of parallel style memory banks and one semantic memory bank. We use contrastive loss based on style banks to achieve intra-domain style invariance. We construct a memory semantic feature bank (``jury'') to achieve inter-domain semantic invariance.}
\label{fig:framework}
\vspace{-0.2in}
\end{figure*}

Generally, a bird's-eye view of our whole framework is illustrated in Fig. {\ref{fig:framework}}: We maintain an encoder to extract features of input images, and use a parallel memory encoder to generate and release memory features in the memory bank.  Within both encoder and memory encoder structure, we both include a style encoder component and semantic encoder component. We explain the usage of these components in the following sections.

\subsection{Intra-domain Invariance on Style Features} \label{sec:intrastyle}
Our first objective is to impose cross-instance intra-domain style invariance via a memory bank construction. The plan is to maintain $D$ domain specific style banks that gradually become agnostic to semantics as the training progresses, while each style bank retains intra-domain invariant and inter-domain contrastive style features at instance level.

We compute the style feature of each $\bx_{d,i}$ via memory encoder: $\bar\bs_{d,i}= E_{m,s}(E_{m,f}(\bx_{d,i}, \btheta_{m,f}), \btheta_{m,s})$, and we sequentially push each arriving $\bar\bs_{d,i}$ into the ``domain style bank'' $\bV^s_d$. We define $D$ number of parallel style memory banks $\bV^s$=($\bV^s_1,...,\bV^s_d,...,\bV^s_D$, $d\in [1,D]$) and dynamically update each $\bV^s_d$ bank like queues: We push each newly arriving style representation $\bar\bs_{d,i}$ into the queue tail of $\bV^s_d$, and remove the oldest style feature from the queue head as in \cite{he2020momentum}. Assuming the memory size of each bank $\bV^s_d$ is a constant $B$. The entries stored in $\bV^s_d$ are denoted with brand new subscripts as: $ \lbrack \bv^s_{d,1},...\bv^s_{d,j},...., \bv^s_{d,B}\rbrack, j\in [1,B]$, where each $\bv^s_{d,j}$ dynamically stores the style feature vector $\bar\bs_{d,\cdot}$ at the $j$ position of the $\bV^s_d$ bank.

Recall that we desire to figure out the shared invariant style features embedded behind each domain. To achieve this goal, we require every style feature out of the style encoder $\bs_{d,i}=E_{s}(E_f(\bx_{d,i}, \btheta_{e,f}), \btheta_{e, s})$ to have high similarity score with any stored style features from the same domain in memory $\bV^s_d$, whereas similarity score between style feature $\bs_{d,i}$ and all features from other domain banks $\bV^s_{d'}, {d'} \neq d$ remains low. We find contrastive loss \cite{he2020momentum,tian2019contrastive} a natural fit to fulfill this goal:
\begin{equation}
\label{eq:styleNCE}
\tiny
\mathcal{L}_s= -  \frac{1}{Z_s} \sum_{d,i,j}
 \log    \frac{\exp( \left \langle \bs_{d,i},  \bv_{d,j}^s \right \rangle / \tau)}{\exp( \left \langle \bs_{d,i},  \bv_{d,j}^s \right \rangle / \tau) + \sum\limits_{d' \neq d} \sum\limits_{\ell=1}^{B}  \exp( \left \langle \bs_{d,i},  \bv_{d',\ell}^{s} \right \rangle / \tau)},
\end{equation}
where $Z_s=B\cdot  \sum_d N_d$ normalizes sample number, $\tau$ is a temperature parameter and $\left \langle \bx_1, \bx_2 \right \rangle  = \bx_1^T \bx_2/ \| \bx_1\| \| \bx_2\|$ denote the cosine similarity between $\bx_1$ and $\bx_2$.

It is transparent that Eq. (\ref{eq:styleNCE}) is essentially a softmax function aiming to distinguish each $( \bs_{d,i},  \bv_{d,j}^s )$ intra-domain pair from the total $B \times (D-1)$ number of $(\bs_{d,i},  \bv_{d',\ell}^{s}), d' \neq d $ inter-domain pairs. As the training proceeds, the instance level style feature $\bs_{d,i}$ compares with all the members in the domain style bank $\bv_{d,j}^s$, $j\in [1,B]$, in order to reach consensus on intra-domain style invariance of domain $d$. In other words, Eq. (\ref{eq:styleNCE}) penalizes feature misalignment between $\bs_{d,i}$ and $\bv_{d,j}^s$, whereas the style features from distinct domains are pushed away, i.e., any increase in inner product $\langle \bs_{d,i},  \bv_{d',\ell}^{s} \rangle, d' \neq d$ increases loss Eq. (\ref{eq:styleNCE}). These $D$ number of domain specific style banks therefore gradually become agnostic to semantics during the training, as each bank $\bV^s_d$ retains intra-domain invariant and inter-domain contrastive style features at instance level.

\subsection{Inter-domain Invariance on Semantic Features}\label{sec:intersemantic}
We have tentative style features at hand. We now turn to our second objective: to learn inter-domain semantic invariant feature via another memory bank. Bear in mind that these semantic features are considered the true causal variables that determine the semantics of samples independent of domains. We define $\bx^{+}_{d,i}$ to be a ``variant'' of each training input $\bx_{d,i}$. We call $\bx^{+}_{d,i}$ a ``variant'' of $\bx_{d,i}$, because it is considered as semantically identical to $\bx_{d,i}$. The sample $\bx^{+}_{d,i}$ is randomly chosen from all possible $D$ domains that either is from the same class of $\bx_{d,i}$, or simply is selected from the augmentation pool (RandAug \cite{cubuk2020randaugment}) of image $\bx_{d,i}$.

Feature $\bc_{d,i}$ is considered the true causal variable that determines the semantics of training samples. We desire that a semantic feature $\bc_{d,i}=E_{c}(  E_f( \bx_{d,i}  , \btheta_{e, f} ), \btheta_{e, c})$ to return a high similarity score with the semantic feature of any possible $\bx^{+}_{d,i}$, whereas $\bc_{d,i}$ remains distinct to samples from other classes. We certainly cannot build $n_c$ classes number of parallel semantic banks analogously to what we did for style banks, because there might be millions of classes. We correspondingly come up with a novel ``jury'' mechanism that relaxes our objective.

{\bf Construction of Semantic Jury Memory Bank}. The semantic ``jury'' bank's construction relies on memory encoders: We sequentially push any arriving semantic memory feature $\bar \bc_{d,i}^+=E_{m,c}( E_{m,f}(\bx^+_{d,i}, \btheta_{m,f}) , \btheta_{m,c})$ into the single semantic feature memory bank $\bV^c$ regardless of whose ``variant'' $\bar \bc_{d,i}^+$ is and which domain $d$ is. We update the entries in $\bV^c$, again, like maintaining a queue structure: entries stored in $\bV^c$ are denoted as: $\bV^c= \lbrack \bv^c_{1},...\bv^c_{j},..., \bv^c_{B}\rbrack, j\in[1,B]$.  Note features $\bc_{d,i}$ and $\bar \bc_{d,i}^+$ are semantically identical, as they represent the different samples from the same class. But how far are $\bc_{d,i}$ and $\bar \bc_{d,i}^+$ in the semantic embedding space?

We leave the judgment to the ``jury''. All the queuing semantic features in the bank have the right to bid, contribute, and weigh their contribution into the similarity measurement between $\bar \bc_{d,i}^+$ and $\bc_{d,i}$. Mathematically, we denote the probability $\bp(\bx_{d,i};\btheta_{e,c}, \btheta_{e,f}, \bV^c)=\lbrack p^e_1,...p^e_j, ..., p^e_{B} \rbrack, j\in[1,B]$ as similarity score between $\bc_{d,i}$ with respect to all stored semantic features in semantic memory $\bV^c$, where each probability entry $p^e_j$ is defined as:
\vspace{-0.05in}
\begin{equation}
\label{eq:p}
\small
p^e_j = \frac{\exp( \left \langle \bc_{d,i},  \bv^{c}_j \right \rangle / \tau)}{ \sum_{\bv^{c} \in \bV^c}  \exp( \left \langle \bc_{d,i},  \bv^c \right \rangle / \tau)     }.
\vspace{-0.05in}
\end{equation}
Similarly, we also compute $\bp(\bx^{+}_{d,i};\btheta_{m,c}, \btheta_{m,f}, \bV^c)=\lbrack p^m_1,..., p^m_{B} \rbrack$, which is the similarity scores between $\bar \bc^+_{d,i}$ with respect to each of the member in the $\bV^c$:
\vspace{-0.05in}
\begin{equation}
\label{eq:pprime}
\small
p^m_j = \frac{\exp( \langle \bar\bc^+_{d,i},  \bv^{c}_j \rangle / \tau)}{ \sum_{\bv^{c} \in \bV^c}  \exp( \left \langle \bar\bc^+_{d,i},  \bv^c \right \rangle / \tau)     }.
\vspace{-0.05in}
\end{equation}

The motivation here is that, the ``jury'' $\bV^c$ would cross check every ``jury'' member's similarity score with both $\bar \bc^+_{d,i}$ and $\bc_{d,i}$. The bank $\bV^c$ summaries these two distribution respectively into $\bp(\bx_{d,i};\btheta_{e,c}, \btheta_{e,f}, \bV^c)$ and $\bp(\bx^{+}_{d,i};\btheta_{m,c}, \btheta_{m,f}, \bV^c)$ as in Eq. (\ref{eq:p}) and Eq. (\ref{eq:pprime}). The assumption here is that, if $\bx^{+}_{d,i}$ and $\bx_{d,i}$ truly share invariant semantic features, then their similarity score across the entire semantic feature bank $\bV^c$ shall be as close as possible, too. We therefore penalize cross entropy between $\bp(\bx^+_{d,i};\btheta_{m,c}, \btheta_{m,f}, \bV^c)$  and $\bp(\bx_{d,i};\btheta_{e,c}, \btheta_{e,f}, \bV^c)$:
\begin{equation}
\scriptsize
\mathcal{L}_c = - \frac{1}{Z_c}\sum_{d,i}  \bp(\bx^+_{d,i};\btheta_{m,c}, \btheta_{m,f}, \bV^c) \log{\bp(\bx_{d,i};\btheta_{e,c}, \btheta_{e,f}, \bV^c)},
\label{eq:jurycheckloss}
\end{equation}
where $Z_c=\sum_d N_d$ normalizes over samples. Eq. (\ref{eq:jurycheckloss}) penalizes misalignment between probability $\bp(\bx_{d,i};\btheta_{e,c}, \btheta_{e,f}, \bV^c)$ and $\bp(\bx^{+}_{d,i};\btheta_{m,c}, \btheta_{m,f}, \bV^c)$ via similarity cross-checks with all the features stored in the $\bV^c$. As a result, features in the current $\bV^c$ jointly vote on distributional similarity between $\bx^{+}_{d,i}$ and $\bx_{d,i}$. This strategy strongly contrasts with conventional contrastive learning, where for MoCo like mechanisms, only a single positive key is considered for each query, whereas negative keys in the memory bank are only considered as negative and contrastive to the positives.

But why not directly penalize the inner product $\langle\bc_{d,i}, \bar \bc^+_{d,i}\rangle$ or the likes which seems a more obvious option? If one takes a closer look at Eq. (\ref{eq:jurycheckloss}), the net effect of the ``jury'' mechanism is that, any feature member, say $\bv_j^c$ in the bank $\bV^c$ having a relatively high similarity score with both $\bc_{d,i}$ and $\bar\bc^+_{d,i}$ would vote for agreement on this similarity, and then $\bv^c_{j}$ becomes confident to contribute itself to jointly supervise the optimization direction to further improve semantic invariance among $\bc_{d,i}$, $\bar\bc^+_{d,i}$ and $\bv_j^c$. The strength of this vote depends on $\bv_j^c$'s similarity score with each $\bc_{d,i}$ and $\bar\bc^+_{d,i}$. This makes one reminiscent of popular multi-view contrastive learning strategy \cite{caron2020unsupervised, tian2019contrastive}, where introducing more views for each instance is always beneficial for learning. Conversely, all feature members in the jury showing low similarity scores with both $\bc_{d,i}$ and $\bar \bc^+_{d,i}$  would automatically tend to contrast itself away from both $\bc_{d,i}$ and $\bar \bc^+_{d,i}$. The loss therefore inherently summarizes the compounding effect of all features in the bank via such joint similarity cross-check mechanism, in comparison to any conventional contrastive learning approaches that banks are only considered ``negative''. We notice a related work \cite{mitrovic2020representation}, that also enforces invariant prediction regularizer across augmentations. We argue that the work in \cite{mitrovic2020representation} is a completely unsupervised algorithm that distinguishes itself from our DG task. Notably, the proposed bank definition here out of the specific DG invariance hypothesis is also orthogonal to \cite{mitrovic2020representation}, and the loss Eq. (\ref{eq:jurycheckloss}) proposed here leads to entirely different interpretation and implementation.

To make sure the features are indeed semantically discriminative, we apply supervised classification loss on $\bx_{d,i}$:
\begin{equation}
\label{eq:sourceentropy}
\mathcal{L}_{cls} =- \frac{1}{N} \sum_{i=1}^{N} {y_{d,i}}\log{C( \bc_{d,i}, \bphi)},
\end{equation}
where classifier $C( \bc_{d,i}, \bphi)$ predicts the class of sample $\bx_{d,i}$ by only using its semantic feature $\bc_{d,i}$.

\subsection{Decoupling Semantics from Styles}
Last but not least, current loss on semantic features $\bc_{d,i}$ is completely disconnected from its style features $\bs_{d,i}$, and cannot benefit from the style invariance assumption at all. A simple solution to fix this is to enforce orthogonality \cite{bousmalis2016domain} between $\bc_{d,i}$ and $\bs_{d,i}$. Let $\bH_c$ and $\bH_s$ be matrices whose rows are the semantic representations $\bc_{d,i}$ and style representations $\bs_{d,i}$ from $\bx_{d,i}$. We constrain the semantic feature to go towards the orthogonal direction against style features, such that the semantic encoder $E_c(\cdot, \btheta_{e, c})$ and classifier $C( \bc_{d,i}, \bphi)$ could mostly avoid overfitting to the style features. This is formalized into the squared Frobenius norm:
\begin{equation}
\mathcal{L}_o = \| \bH^{T}_c \bH_s \|^{2}_F.
\label{eq:orthogonal}
\end{equation}

\subsection{Overall Loss Function}
Taking into account all the discussions above, the eventual loss function is:
\begin{equation}
\mathcal{L}=\mathcal{L}_{cls}+ \mathcal{L}_{s}+ \mathcal{L}_{c}+  \mathcal{L}_{o},
\label{eq:finalloss}
\end{equation}
where all present losses are equally weighted. The overall loss is backpropagated through the network in a batchwise manner. The parameters of encoder $\bTheta_{e}=\{\btheta_{e, f},\btheta_{e, c},\btheta_{e, s}\}$ and classifier $\bphi$ are updated via backpropagation, whereas the parameters of memory encoder $\bTheta_m=\{\btheta_{m, f},\btheta_{m, c},\btheta_{m, s}\}$ are momentum updated~as:
\begin{equation}
\bTheta_m = \alpha \times \bTheta_m + (1-\alpha) \times \bTheta_e,
\end{equation}
where  $\alpha \in \left[0, 1 \right)$ is a momentum coefficient. We only use obtained encoder parameters $\btheta_{e, f},\btheta_{e, c}$ and classifier parameters $\bphi$ during test time inference.

\subsection{Extension to MSDA}\label{sec:msda}
Another advantage of the proposal is that, loss Eq. (\ref{eq:finalloss}) can be transferred to deal with multi-source domain adaptation (MSDA) tasks with few modifications. The only difference is that for target data without any annotations, only those augmentation images of $\bx_{d,i}$ are used to define $\bx^+_{d,i}$ and there is no classifier applied on target data semantic features, as the class label is not available. Except for this difference, all target data can be conveniently plugged into our algorithm with a domain ID, $d \in [1,D]$. The training procedure then progresses under the exactly same losses and network constructions as we did for DG tasks.

\begin{table*}[t]
  \centering
  \scriptsize
  \caption{Performance (\%) comparisons with the state-of-the-art approaches for DG.}
  \setlength{\tabcolsep}{1.5mm}{
    \begin{tabular}{l| cccc|c||cccc|c||cccc|c}
    \hline
    \multicolumn{1}{c|}{\multirow{2}[1]{*}{Method}} & \multicolumn{5}{c||}{Digits-DG}           &\multicolumn{5}{c||}{PACS} &\multicolumn{5}{c}{Office-Home}  \\
\cline{2-16}   & MNIST & MNIST-M & SVHN & SYN  & Avg   & Art  & Cartoon   & Photo   & Sketch   & Avg & Artistic  & Clipart   & Product   & Real World   & Avg \\
    \hline
    MMD-AAE  \cite{li2018domain}  & 96.5  & 58.4  & 65.0  & 78.4  & 74.6 & 75.2  & 72.7  & 96.0  & 64.2  & 77.0  & 56.5 & 47.3 & 72.1 & 74.8 & 62.7  \\
    CCSA  \cite{motiian2017unified} & 95.2 &58.2 & 65.5  & 79.1  & 74.5 & 80.5 & 76.9  & 93.6  & 66.8  & 79.4 &59.9 & 49.9 & 74.1 & 75.7 & 64.9  \\
    JiGen  \cite{carlucci2019domain} & 96.5  & 61.4  & 63.7  & 74.0  & 73.9 & 79.4  & 75.3  & 96.0 & 71.6  & 80.5 & 53.0 & 47.5 & 71.5 & 72.8 & 61.2   \\
    CrossGrad  \cite{shankar2018generalizing} & 96.7  & 61.1  & 65.3  & 80.2 & 75.8 & 79.8  & 76.8  & 96.0  & 70.2  & 80.7 & 58.4 & 49.4 & 73.9 & 75.8 & 64.4   \\
    Epi-FCR  \cite{li2019episodic} & -     & -     & -     & -  & - & 82.1 & 77.0 & 93.9  & 73.0 & 81.5 & -     & -     & -     & -  & -  \\
     EISNet ~\cite{wang2020learning} & -     & -     & -     & -  & - & 81.9 & 76.4 & 95.9  & 74.3 & 82.1 & -     & -     & -     & -  & -  \\
    L2A-OT \cite{zhou2020learning} & 96.7     & 63.9     & 68.6     & 83.2  & 78.1 & 83.3 & 78.2 & 96.2  & 73.6 & 82.8 & 60.6 & 50.1 & 74.8 & 77.0 & 65.6  \\
    DecAug \cite{bai2020decaug} & -     & -     & -     & -  & - & 79.0 & 79.6 & 95.3  & 75.6 & 82.4 & -     & -     & -     & -  & -  \\
    MixStyle \cite{zhou2021domain}  & 96.5     & 63.5     & 64.7     & 81.2  & 76.5 & 84.1 & 78.8 & 96.1  & 75.9 & 83.7 & 58.7 & \textbf{53.4} & 74.2 & 75.9 & 65.5  \\
    \hline
    Vanilla &95.8   &58.8  &61.7  &78.6  &73.7 & 77.0  & 75.9  & 96.0 & 69.2  & 79.5 & 58.9 & 49.4 & 74.3 & 76.2 & 64.7   \\
     STEAM& \textbf{96.8}  & \textbf{67.5} & \textbf{76.0}  & \textbf{92.2} & \textbf{83.1} & \textbf{85.5}& \textbf{80.6} & \textbf{97.5} & \textbf{82.9}  & \textbf{86.6}  & \textbf{62.1} & 52.3 & \textbf{75.4} & \textbf{77.5} & \textbf{66.8}\\
    \hline
    \end{tabular}}%
  \label{tab:digitsandpacs_dg}%
  \vspace{-4mm}
\end{table*}%

\section{Experiments}

\subsection{Evaluation on Domain Generalization}
\textbf{Datasets.} We perform DG tasks via extensive evaluations on the following benchmarks: (1) \textbf{Digits-DG} \cite{zhou2020learning} includes $4$ domains (\textit{MNIST} \cite{lecun1998gradient}, \textit{MNIST-M} \cite{ganin2015unsupervised}, \textit{SVHN} \cite{netzer2011reading} and \textit{SYN} \cite{ganin2015unsupervised}) with an evident domain shift in font style, stroke color and background. (2) \textbf{PACS} \cite{li2017deeper} is a widely used domain generalization benchmark, which is composed of four domains (\textit{Art Painting}, \textit{Cartoon}, \textit{Photo} and \textit{Sketch}). Each domain includes samples from $7$ different categories, including a total of $9,991$ samples. (3) \textbf{Office-Home} \cite{venkateswara2017deep} contains around $15,500$ images of $65$ classes, distributed across $4$ domains (\textit{Artistic}, \textit{Clipart}, \textit{Product} and \textit{Real world}). (4) \textbf{DomainNet} \cite{peng2019moment} is a recently established large-scale dataset for multi-source domain adaptation and domain generalization, which includes about $0.6$ million images in $345$ classes distributed across $6$ domains (i.e., \textit{Clipart}, \textit{Infograph}, \textit{Quickdraw}, \textit{Painting}, \textit{Real}, \textit{Sketch}).

For a fair comparison with prior works, we follow the standard leave-one-domain-out evaluation procedure as in \cite{carlucci2019domain, li2017deeper, li2019episodic}, where one domain is chosen as the unseen target and the remaining domains are used as source domains for training. For Digits-DG, PACS and Office-Home, the authors of \cite{zhou2020learning, zhou2020domain, zhou2021domain} have specified particular train and val splits for each domain to ensure a fair comparison. They use the entire \textit{train} $+$ \textit{val} target data as the \textit{test} data. We use the same data split definition for our experiments. For DomainNet, according to \cite{chattopadhyay2020learning}, we employ their dataset division and report the accuracy on the $\textit{test}$ split of target domain.

\textbf{Implementation details.} Following the backbone setting of \cite{zhou2020learning}, we use $4$ \textit{conv} layers and a softmax layer for Digits-DG dataset. ReLU and $2\times2$ max-pooling are inserted after each convolution layer. The model is trained with SGD, initial learning rate of $0.05$ and batch size of $128$ for $50$ epochs. For both PACS and Office-Home, we use ResNet-18 pretrained on ImageNet as the CNN backbone, as in \cite{zhou2020learning, zhou2021domain}. We train the model with SGD, initial learning rate of $0.002$ and batch size of $30$ for $60$ epochs. The learning rate is further decayed by the cosine annealing rule. For DomainNet, we experiment with ResNet-18 and ResNet-50 backbone architectures, as in \cite{chattopadhyay2020learning}. For all experiments, the semantic encoder, style encoder and classifier are all implemented using a fully connected (FC) layer. The size of style and semantic memory bank is set as $2,048$.

\textbf{Baselines.} To evaluate our method, we consider comparisons with the following approaches: (1) \textbf{Vanilla} simply trains the plain classification model on all available source domains using all annotations, the model is then directly used to classify target samples. (2) \textbf{CrossGrad} \cite{shankar2018generalizing} perturbs input using adversarial gradients from a domain classifier. (3) \textbf{CCSA} \cite{motiian2017unified} explores a contrastive semantic alignment loss for domain-invariant representation learning. (4) \textbf{MMD-AAE} \cite{li2018domain} imposes an MMD loss on the hidden layers of an autoencoder. (5) \textbf{JiGen} \cite{carlucci2019domain} utilizes an auxiliary self-supervision loss so that the features can be used to solve the Jigsaw puzzle task. (6) \textbf{Epi-FCR} \cite{li2019episodic} designs an episodic training strategy. (7) \textbf{EISNet} \cite{wang2020learning} develops a momentum metric learning scheme with the $K$-hard negative mining to improve the network generalization ability.  (8) \textbf{L2A-OT} \cite{zhou2020learning} synthesizes extra data from pseudo-novel domains to augment the source domains. (9) \textbf{DecAug} \cite{bai2020decaug} generates extra data augmentations through perturbing the disentangled style feature and semantic features. (10) \textbf{DMG} \cite{chattopadhyay2020learning} learns domain specific masks for generalization on different domains. (11) \textbf{MixStyle} \cite{zhou2021domain} mixes instance level feature statistics of training samples across various sources to introduce more domain diversity via synthesizing.

\textbf{Results on Digits-DG.} Table \ref{tab:digitsandpacs_dg} shows that, our method exhibits clear advantages over the existing state-of-the-art methods. Please note that our model is especially effective and commanding on challenging DG directions, e.g., \textit{MNIST-M} and \textit{SVHN}, as they seem to have large domain variations compared with other directions. This is a valid justification that the instance level style invariance along with the proposed ``jury'' mechanism together is a legitimate idea to deal with domain generalization problems. Compared with the methods that hinge on marginal distribution alignment across domains, e.g., MMD-AAE and CCSA, our model even demonstrates around $8.5\%$ performance boost on average. This well validates our assumption in Section \ref{sec:intrastyle}, that intra-domain style invariance seems to be a more practical and suitable hypothesis in presence of computing background statistics hidden in domain styles. In other words, the prior knowledge on intra-domain style invariance effectively reduce the uncertainty when searching for optimal network parameters so that the parameters reduces overfitting to domain styles.

\begin{table}[t]
\center
\scriptsize
\setlength{\tabcolsep}{4pt}
\renewcommand{\arraystretch}{1.3}
\caption{Leave-one-domain-out results on DomainNet for DG.}
\vspace{0.2cm}
\begin{tabular}{l l l  c  c c c c c c | c }
\toprule
& & \textbf{Method} & & Clp & Inf & Pnt & Qdr & Rel & Skt & Avg. \\
\midrule
\multirow{6}{*}{\rotatebox{90}{\centering ResNet-18 }}
& & Vanilla  && 56.5 & 18.4 & 45.3 & 12.4 & 57.9 & 38.8 & 38.2\\
& & Multi-Headed ~\cite{chattopadhyay2020learning} && 55.4 & 17.5 & 40.8 & 11.2 & 52.9 & 38.6 & 36.1\\
& & MetaReg~\cite{balaji2018metareg} && 53.6 & 21.0 & 45.2 & 10.6 & 58.4 & 42.3 & 38.5\\
& & DMG ~\cite{chattopadhyay2020learning} && \textbf{60.0} & 18.7 & 44.5 & 14.1 & 54.7 & 41.7 & 39.0 \\
\cline{3-11}
& & STEAM  && 58.3 & \textbf{22.1} & \textbf{47.4} & \textbf{14.4} & \textbf{58.6} &\textbf{45.9} & \textbf{41.1} \\

\midrule
\multirow{6}{*}{\rotatebox{90}{\centering ResNet-50 }}
& & Vanilla  && 64.0 & 23.6 & 51.0 & 13.1 & 64.4 & 47.7 & 44.0\\
& & Multi-Headed ~\cite{chattopadhyay2020learning} && 61.7 & 21.2 & 46.8 & 13.8 & 58.4 & 45.4 & 41.2\\
& & MetaReg~\cite{balaji2018metareg} && 59.7 & 25.5 & 50.1 & 11.5 & 64.5 & 50.0 & 43.6\\
& & DMG~\cite{chattopadhyay2020learning}  && \textbf{65.2} & 22.1 & 50.0 & 15.6 & 59.6 & 49.0 & 43.6\\
\cline{3-11}
& & STEAM  && 64.6 & \textbf{27.0} & \textbf{54.0} & \textbf{15.8} & \textbf{65.6} & \textbf{52.2} & \textbf{46.5}\\
\bottomrule
\end{tabular}
\vspace{-0.6cm}
\label{tab:domainnet-dg}
\end{table}

\textbf{Results on PACS.}
This part of results is shown in Table \ref{tab:digitsandpacs_dg}. Our method achieves the best performance on all test domains. Note the recently proposed EISNet also involves the usage of a feature memory. However, the memory in EISNet is only used for the sake of hard triplets selection without consideration on feature invariance.  In DecAug \cite{bai2020decaug}, similar semantic-style orthogonal regularization loss was used. However, orthogonality is perhaps the most widely used tool everywhere, and we simply employ orthogonality as an auxiliary regularizer. Nevertheless, our motivation and hypothesis are completely different from DecAug, highlighting the outstanding performance given our instance-level style invariance and ``jury'' mechanism.

\textbf{Results on Office-Home.}
The results are shown in Table \ref{tab:digitsandpacs_dg}. Our STEAM achieves the best average performance. Notably, the simple vanilla model shows strong results on this benchmark. Most of baselines provide only marginal improvements than vanilla model that are under $1.0\%$. As discussed in L2A-OT, this might be owing to the fact that Office-Home is relatively a large composition of data, compared with PACS and Digits-DG, thus offering inherently bigger domain diversity in training data already. In contrast, our method shows the best performance on average with an impressive $2.1\%$ improvement over the Vanilla.

\textbf{Results on DomainNet.}
DomainNet is considered as perhaps the most challenging benchmark, owing to its dataset size, both in terms of image number and category numbers. Table \ref{tab:domainnet-dg} shows that, on DomainNet dataset, the Vanilla baseline achieves competitive results in comparison to domain generalization methods MetaReg and DMG, while our method again surpasses all competitors. We observe that our model is leading in the table with improved average performance, especially when ResNet-18 and ResNet-50 are used as the backbone architectures, showing respectively $2.1\%$ and $2.5\%$ accuracy improvement. At this point, it is worthy to mention that our method has achieved better robustness, for consistently offering better performance than other baselines.

\subsection{Evaluation on Domain Adaptation}
\textbf{Datasets and implementation details.} To justify STEAM's validity on the application of MSDA, we implement STEAM under the problem definition described in Section \ref{sec:msda}. We firstly consider evaluation on PACS dataset again with the same problem definition and setting of \cite{yang2020curriculum}. Next, we follow \cite{zhou2020domain} and use miniDomainNet for evaluation, which is a sampled subset of DomainNet and reformatted into a smaller image size ($96\times96$). In general, miniDomainNet consists of $4$ domains (\textit{Clipart}, \textit{Painting}, \textit{Real} and \textit{Sketch}) across $126$ classes, which mostly resembles data diversity of the original DomainNet. For PACS, we use ResNet-18 pretrained on ImageNet as the CNN backbone, by following training protocols in \cite{yang2020curriculum}. Batch size is $32$. For miniDomainNet, we use ResNet-18 as the CNN backbone, the same used as in \cite{zhou2020domain}. Similarly, we use SGD with momentum as the optimizer, and the learning rate decays according to cosine annealing rule. The model is trained with an initial learning rate of $0.005$ for $60$ epochs. For each mini-batch, we sample from each domain $64$ images.

\begin{table}[t]
\center
\footnotesize
\renewcommand{\arraystretch}{1.1}
\caption{Domain adaptation results on PACS.}
\vspace{0.2cm}
\begin{tabular}{l | c c c c | c}
\hline
Method & Art & Cartoon & Photo & Sketch & Avg \\
\hline
Source-only & 74.9 & 72.1 & 94.5 & 64.7 & 76.6 \\
DANN \cite{ganin2015unsupervised} & 81.9 & 77.5 & 91.8 & 74.6 & 81.5 \\
MDAN \cite{zhao2018adversarial} &79.1 & 76.0 & 91.4 & 72.0 & 79.6 \\
WBN \cite{mancini2018boosting} & 89.9 & 89.7 & 97.4 & 58.0 & 83.8 \\
MCD \cite{saito2018maximum} &88.7 & 88.9 & 96.4 & 73.9 & 87.0 \\
M$^3$SDA \cite{peng2019moment} & 89.3 & 89.9 & 97.3 &76.7 & 88.3 \\
CMSS \cite{yang2020curriculum} & 88.6 & 90.4 & 96.9 & 82.0 & 89.5 \\
\hline
STEAM & \textbf{94.0} & \textbf{93.7} & \textbf{99.3} & \textbf{85.1} & \textbf{93.0} \\
\hline
\end{tabular}
\vspace{-0.1in}
\label{tab:pacs-da}
\end{table}%

\begin{table}[t]
\center
\footnotesize
\renewcommand{\arraystretch}{1.1}
\caption{Domain adaptation results on miniDomianNet.}
\vspace{0.2cm}
\begin{tabular}{l | c c c c | c}
\hline
Method & Clipart & Painting & Real & Sketch & Avg \\
\hline
Source-only & 63.4 & 49.9 & 61.5 & 44.1 & 54.7 \\
MCD \cite{saito2018maximum} & 62.9 & 45.7 & 57.5 & 45.8 & 53.0 \\
DCTN~\cite{xu2018deep} &62.0 & 48.7 & 58.8 & 48.2 & 54.4 \\
DANN \cite{ganin2015unsupervised} & 65.5 & 46.2 & 58.6 & 47.8 & 54.6 \\
M$^3$SDA \cite{peng2019moment} & 64.1 & 49.0 & 57.7 & 49.2 & 55.0 \\
MME \cite{saito2019semi} & 68.0 & 47.1 & 63.3 & 43.5 & 55.5\\
DAEL \cite{zhou2020domain} & 69.9 & 55.1 & 66.1 &55.7 & 61.7 \\
\hline
STEAM & \textbf{71.4} & \textbf{61.9} & \textbf{71.1} & \textbf{60.9} & \textbf{66.3} \\
\hline
\end{tabular}
\vspace{-0.4cm}
\label{tab:minidomainnet-da}
\end{table}%

\textbf{Results.} Given the standard test protocol in \cite{peng2019moment} on PACS, we use one domain as target and the remaining as sources. Classification accuracy on the target domain test set is reported. We compare our STEAM with two state-of-the-art multi-source domain adaptation approaches: M$^3$SDA \cite{peng2019moment} and CMSS \cite{yang2020curriculum}. In addition, we also present the following methods as our baselines: DANN \cite{ganin2015unsupervised}, MDAN \cite{zhao2018adversarial}, WBN \cite{mancini2018boosting}, MCD \cite{saito2018maximum}. The results are shown in Table \ref{tab:pacs-da}. Our method achieves the state-of-the-art average accuracy of $93.0\%$. On the most challenging \textit{sketch} domain, we obtain $85.1\%$, clearly outperforming other baselines. On the miniDomainNet, we compare with the same methods presented in DAEL \cite{zhou2020domain}, including MCD \cite{saito2018maximum}, DCTN \cite{xu2018deep}, DANN \cite{ganin2015unsupervised}, M$^3$SDA \cite{peng2019moment}, MME \cite{saito2019semi}. The results are shown in Table \ref{tab:minidomainnet-da}. Our method achieves $66.3\%$ average accuracy, again, justifying significant advantages out of our interesting invariance hypothesis.

\begin{table}[t]
\center
\scriptsize
\setlength{\tabcolsep}{4pt}
\renewcommand{\arraystretch}{1.4}
\caption{Ablation on Digits-DG. MT, MM, SV, and SY indicates \textit{MNIST}, \textit{MNIST-M}, \textit{SVHN}, and \textit{SYN}, respectively.}
\vspace{0.2cm}
\begin{tabular}{l | c c c c  |c c c c | c}
\hline

Method &$\mathcal{L}_{cls}$ &$\mathcal{L}_{s}$ &$\mathcal{L}_{o}$ &$\mathcal{L}_{c}$ & MT & MM & SV & SY & Avg. \\ \hline
Vanilla &$\surd$ &~ &~ &~ & 95.8 & 58.8 & 61.7 & 78.6 & 73.7 \\
Vanilla-style &$\surd$ &$\surd$ &$\surd$ &~ & 96.3 & 63.6 & 69.3 & 82.4 & 77.9 \\
Vanilla-semantic &$\surd$ &~ &~ &$\surd$ & 96.0 & 65.2 & 74.5 & 86.2 & 80.5 \\
\hline
STEAM &$\surd$ &$\surd$ &$\surd$ &$\surd$ & \textbf{96.5} & \textbf{67.5} & \textbf{76.0} & \textbf{92.2} & \textbf{83.0} \\
\hline
\end{tabular}
\vspace{-0.18in}
\label{tab:ablation-digits-dg}
\end{table}

\subsection{Further Analysis}
\textbf{Ablation study.} We investigate the impact of each component in Eq. (\ref{eq:finalloss}) by comparing several variations of STEAM using the Digits-DG and PACS datasets. \textbf{Vanilla:} a conventional supervised learning formulation that uses all source domains for training, i.e., with training loss $\mathcal{L}_{cls}$; \textbf{Vanilla-style:} we add intra-domain style invariance loss and orthogonal regularization to Vanilla model, i.e., train using loss $\mathcal{L}_{cls}+ \mathcal{L}_{s} + \mathcal{L}_o$; \textbf{Vanilla-semantic:} we only add inter-domain invariance constraint on semantic features, i.e., training loss $\mathcal{L}_{cls}+ \mathcal{L}_c$; \textbf{STEAM:} our complete training scheme, with training loss defined in Eq. (\ref{eq:finalloss}). Note we don't separate the $\mathcal{L}_s$ and  $\mathcal{L}_o$ here, but please see supplementary file for more related ablation studies.

Table \ref{tab:ablation-digits-dg} and \ref{tab:ablation-pacs} display the results. Notably, we observe three phenomenons: (1) \emph{Vanilla-style} outperforms \emph{Vanilla} by average $4.2\%$ and $3.5\%$ on Digits-DG and PACS, supporting the advantage of ``intra-domain style invariance'' hypothesis for domain generalization alone; (2) \emph{Vanilla-semantic} also surpasses over \emph{Vanilla} by $6.8\%$ on Digits-DG, and $4.9\%$ on PACS, which well validates the effectiveness of semantic invariance imposed through ``jury'' mechanism alone. (3) By leveraging on both intra-domain style invariance and inter-domain semantic invariance assumptions, our entire STEAM framework further drastically improves the performance across all settings, indicating the complementary roles of these two factors.

\textbf{Design choices for STEAM.} We now explore alternative definitions for each loss term, to better understand the essence behind STEAM. (1) \textbf{Domain classification:} we replace the entire style bank and associated instance level contrastive loss described in Eq. (\ref{eq:styleNCE}) by a domain classifier. We then use the domain classifier to predict the domain label, so that the style features becomes domain-discrinimative. (2) \textbf{L2-matching:} we replace Eq. (\ref{eq:jurycheckloss}), by directly minimizing the $l2$-distance of each semantic feature pairs $\|\bc_{d,i}- \bar \bc^+_{d,i}\|^2_2$ where $\bc_{d,i}$ and $ \bar \bc^+_{d,i}$ share identical class label. (3) \textbf{Contrastive:} we define semantic features $\bar \bc^+_{d,i}$ as the positive key of $\bc_{d,i}$, with random instances sampled from the memory encoders forming  $\bV^c$ as negative samples. We then perform conventional instance level Info-NCE loss on $(\bc_{d,i}, \bar \bc^+_{d,i})$.

\begin{table}[t]
\center
\scriptsize
\setlength{\tabcolsep}{4pt}
\renewcommand{\arraystretch}{1.4}
\caption{Ablation on PACS for domain generaliztion. A, C, P, and S indicates \textit{Art}, \textit{Cartoon}, \textit{Photo}, and \textit{Sketch}, respectively.}
\vspace{0.2cm}
\begin{tabular}{l | c c c c | c c c c | c}
\hline

Method &$\mathcal{L}_{cls}$ &$\mathcal{L}_{s}$ &$\mathcal{L}_{o}$ &$\mathcal{L}_{c}$ & A & C & P & S & Avg. \\ \hline
Vanilla &$\surd$ &~ &~ &~ & 77.0 & 75.9 & 96.1 & 69.2 & 79.5 \\
Vanilla-style &$\surd$ &$\surd$ &$\surd$ &~ & 80.2 & 78.7 & 96.5 & 76.6 & 83.0  \\
Vanilla-semantic &$\surd$ &~ &~ &$\surd$  & 83.3 & 77.6 & 96.7 & 80.1 & 84.4 \\
\hline
STEAM &$\surd$ &$\surd$ &$\surd$ &$\surd$  & \textbf{85.5} & \textbf{80.6} & \textbf{97.5} & \textbf{82.9} & \textbf{86.6} \\
\hline
\end{tabular}
\vspace{-0.15in}
\label{tab:ablation-pacs}
\end{table}

\begin{figure}[t]
\center
\includegraphics[width=0.8\linewidth]{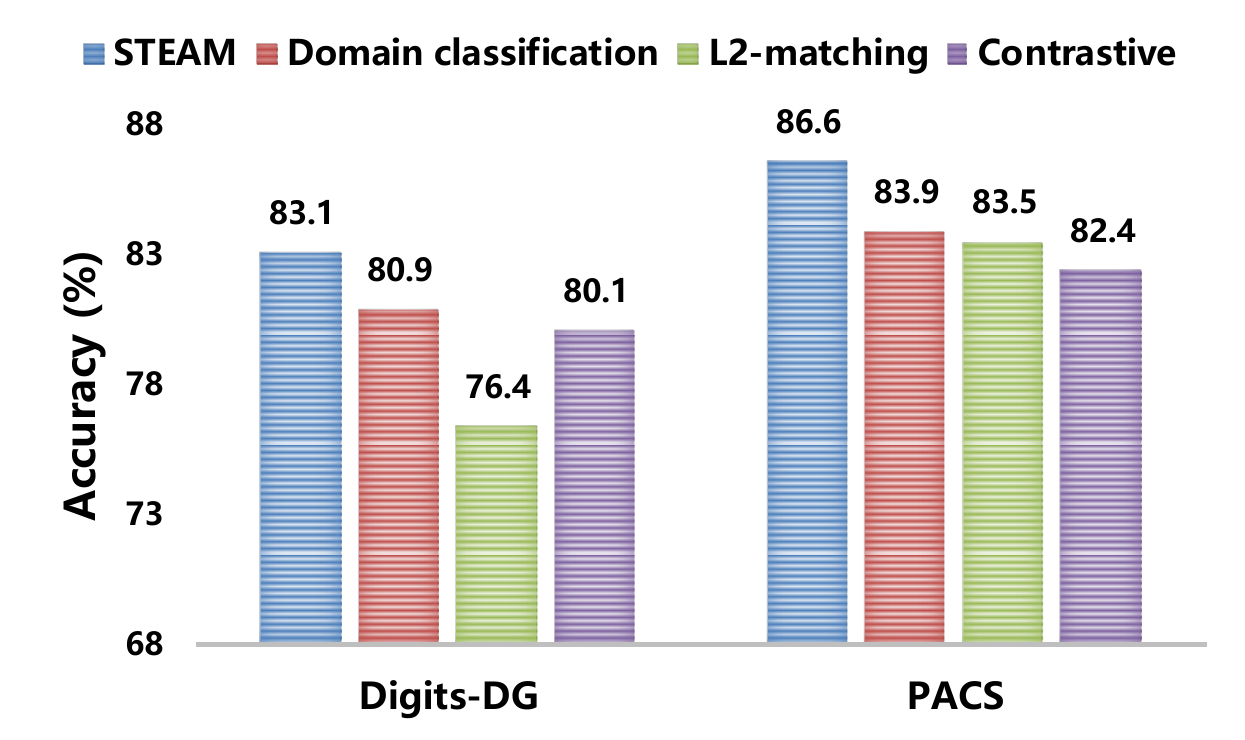}
\caption{Study on different design choices of STEAM.}
\vspace{-0.2in}
\label{fig:design}
\end{figure}

Fig. \ref{fig:design} shows that NONE of the above replacements using alternative losses have achieved any better design than our original STEAM framework. The \emph{Domain classification} achieves some descent test accuracy though (worse than STEAM), while apparently the idea of instance level style invariance is a much stronger prior than simply requiring the style features to be linear separable. \emph{L2-matching} performs worse than our STEAM, meaning directly minimizing the geometry distance within each $(\bc_{d,i}, \bar \bc^+_{d,i})$ pair has ignored important information to be inter-semantically contrastive. Finally, if we would use plain Info-NCE loss as in contrastive learning, i.e., \emph{Contrastive}, worse performance is observed, showing the superiority of our proposed ``jury'' mechanism for inter-domain semantic invariance learning.

\section{Conclusion}
In this paper, we propose a novel algorithm capitalizing on both Style and sEmAntic memory mechanism (STEAM) for domain generalization tasks. Importantly, we find leveraging on intra-domain style invariance can lead to a significant improvement on the efficacy of domain generalization. The intra-domain style invariance prior can help improve the learning of semantic features, owing to reduced overfitting to domain styles during training. We introduce efficient memory bank construction policies for both style and semantic features that store useful statistics for computing our losses. Specifically, our semantic feature bank serves as a ``jury'' and helps effectively improve intra-class invariance cross different domains. Empirical results verify our assumption on various benchmarks.

\textbf{Acknowledgments.} This work was funded by JD AI RESEARCH, NSFC No. 61872329 and the Fundamental Research Funds for the Central Universities under contract WK3490000005.

{\small
\bibliographystyle{ieee_fullname}
\bibliography{egbib}
}

\end{document}